\documentclass[11pt]{article}
\usepackage{acl2017}
\usepackage{times}
\usepackage{url}
\usepackage{soul}
\usepackage{latexsym}
\usepackage{amssymb}
\usepackage{amsmath}
\usepackage{stmaryrd}
\usepackage{color}
\usepackage{graphicx}
\usepackage{comment}
\usepackage{verbatim}
\usepackage{xcolor}
\usepackage{pgfplots}

\DeclareRobustCommand{\hlgreen}[1]{{\sethlcolor{lime}\hl{#1}}}
\DeclareRobustCommand{\hlred}[1]{{\sethlcolor{pink}\hl{#1}}}
\DeclareRobustCommand{\hlyellow}[1]{{\sethlcolor{yellow}\hl{#1}}}

\usepackage{array}
\newcolumntype{L}[1]{>{\raggedright\let\newline\\\arraybackslash\hspace{0pt}}m{#1}}
\newcolumntype{C}[1]{>{\centering\let\newline\\\arraybackslash\hspace{0pt}}m{#1}}
\newcolumntype{R}[1]{>{\raggedleft\let\newline\\\arraybackslash\hspace{0pt}}m{#1}}

\newcommand{\T}{\textit}

\usepackage{xargs} 
\usepackage[colorinlistoftodos,prependcaption,textsize=tiny]{todonotes}
\usepackage{marginnote}
\newcommandx{\todoij}[2][1=]{\todo[inline]{JP: #2}\xspace}
\newcommandx{\todoir}[2][1=]{\todo[inline]{SR: #2}\xspace}
\newcommandx{\todoil}[2][1=]{\todo[inline]{ML: #2}\xspace}

\newcommandx{\todoj}[2][1=]{\todo[linecolor=blue,backgroundcolor=blue!10,bordercolor=blue,#1]{JP: #2}\xspace}
\newcommandx{\todor}[2][1=]{\todo[linecolor=red,backgroundcolor=red!25,bordercolor=red,#1]{SR: #2}\xspace}
\newcommandx{\todol}[2][1=]{\todo[linecolor=cyan,backgroundcolor=cyan!25,bordercolor=cyan,#1]{ML: #2}\xspace}

\usepackage{xspace}
\newcommand \webq{\textsc{Web\-Questions}\xspace}
\newcommand \graphq{\textsc{Graph\-Questions}\xspace}
\newcommand \spades{\textsc{Spades}\xspace}
\newcommand \geoq{\textsc{GeoQuery}\xspace}

\newenvironment{myquote}[1]%
{\list{}{\leftmargin=#1\rightmargin=#1}\item[]}%
{\endlist}

\aclfinalcopy 
\def\aclpaperid{219} 
\title{Learning Structured Natural Language Representations \\for Semantic Parsing}

\author{Jianpeng Cheng$^\dagger$\quad Siva Reddy$^\dagger$\quad Vijay Saraswat$^\ddagger$ \and Mirella Lapata$^\dagger$\\
	$^\dagger$School of Informatics, University of Edinburgh\\
	$^\ddagger$IBM T.J. Watson Research\\
	{\tt \{jianpeng.cheng,siva.reddy\}@ed.ac.uk, vsaraswa@us.ibm.com,} \\
	{\tt mlap@inf.ed.ac.uk}
}

\date{}

\begin{document}
\maketitle
\begin{abstract}
  We introduce a neural semantic parser which converts natural language
  utterances to intermediate
  representations in the form of predicate-argument structures, which
  are induced with a
  transition system and subsequently mapped to target domains. 
  The semantic parser is trained end-to-end using annotated
  logical forms or their denotations. We achieve
  the state of the art on \spades and \graphq and obtain competitive
  results on \geoq and \webq.  The induced predicate-argument
  structures shed light on the types of representations useful for semantic
  parsing and how these are different from linguistically motivated
  ones.\footnote{Our code will be available at \url{https://github.com/cheng6076/scanner}.}

\end{abstract}

\section{Introduction}

Semantic parsing is the task of mapping natural language utterances to
machine interpretable meaning representations. Despite differences in the choice of meaning
representation and model structure, most existing work conceptualizes
semantic parsing following two main approaches. Under the first
approach, an utterance is parsed and grounded to a meaning
representation \emph{directly} via learning a task-specific grammar
\cite{zelle1996learning,zettlemoyer_learning_2005,wong_learning_2006,kwiatkowksi2010inducing,liang2011learning,berant-EtAl:2013:EMNLP,flanigan_discriminative_2014,pasupat_compositional_2015,groschwitz_graph_2015}. Under
the second approach, the utterance is first parsed to an
\emph{intermediate} task-independent representation tied to
a syntactic parser and then mapped to a grounded representation
\cite{kwiatkowski2013scaling,reddy2016transforming,reddy2014large,krishnamurthy_learning_2015,gardner_openvocabulary_2017}. 
A merit of the two-stage approach is that it creates reusable intermediate interpretations,
which potentially enables the handling of unseen words and knowledge
transfer across domains \cite{bender2015layers}.

The successful application of encoder-decoder models
\cite{bahdanau2014neural,sutskever2014sequence} to a variety of NLP
tasks has provided strong impetus to treat semantic parsing as a
sequence transduction problem where an utterance is mapped to a
target meaning representation in string format
\cite{dong2016language,jia2016data,kovcisky2016semantic}.  Such models
still fall under the first approach, however, in contrast to previous
work
\cite{zelle1996learning,zettlemoyer_learning_2005,liang2011learning}
they reduce the need for domain-specific assumptions, grammar
learning, and more generally extensive feature engineering. But this
modeling flexibility comes at a cost since it is no longer possible to
interpret how meaning composition is performed. Such knowledge plays a
critical role in understand modeling limitations so as to build better
semantic parsers. Moreover, without any task-specific prior knowledge,
the learning problem is fairly unconstrained, both in terms of the
possible derivations to consider and in terms of the
target output which can be ill-formed (e.g.,~with extra or missing
brackets).

In this work, we propose a neural semantic parser that alleviates the
aforementioned problems. Our model falls under the second class of
approaches where utterances are first mapped to an intermediate
representation containing natural language predicates. However, rather
than using an external parser
\cite{reddy2014large,reddy2016transforming} or manually specified CCG
grammars \cite{kwiatkowski2013scaling}, we induce intermediate
representations in the form of predicate-argument structures from
data.  This is achieved with a transition-based approach which by
design yields recursive semantic structures, avoiding the problem of generating
ill-formed meaning representations.  Compared to existing chart-based
semantic parsers
\cite{krishnamurthy2012weakly,cai2013large,berant-EtAl:2013:EMNLP,berant2014semantic},
the transition-based approach does not require feature
decomposition over structures and thereby enables the exploration of
rich, non-local features.  The output of the transition system is then
grounded (e.g.,~to a knowledge base) with a neural mapping model under
the assumption that grounded and ungrounded structures are
isomorphic.\footnote{We discuss the merits and limitations of this
  assumption in Section \ref{sec:conclusions}.} As a result, we obtain
a neural network that jointly learns to parse natural language
semantics and induce a lexicon that helps grounding.

The whole network is trained end-to-end on natural language utterances paired with
annotated logical forms or their denotations.  We conduct
experiments on four datasets, including \geoq (which has logical
forms; \citealt{zelle1996learning}), \spades
\cite{bisk2016evaluating}, \webq \cite{berant-EtAl:2013:EMNLP}, and
\graphq \cite{su2016generating} (which have denotations).  Our
semantic parser achieves the state of the art on \spades and
\graphq, while obtaining competitive results on
\geoq and \webq.  A side-product of our modeling framework is that the
induced intermediate representations can contribute to rationalizing
neural predictions \cite{lei2016rationalizing}.  Specifically, they
can shed light on the kinds of representations (especially predicates) useful for semantic
parsing.  Evaluation of the induced predicate-argument relations
against syntax-based ones reveals that they are interpretable and
meaningful compared to heuristic baselines, but they sometimes deviate
from linguistic conventions.

\section{Preliminaries}
\label{sec:preliminaries}

\paragraph{Problem Formulation} Let~$\mathcal{K}$ denote a knowledge
base or more generally a reasoning system, and $x$~an utterance paired with a grounded
meaning representation $G$ or its denotation~$y$. Our problem is to
learn a semantic parser that maps $x$ to~$G$ via
an intermediate ungrounded representation~$U$.  When $G$ is executed
against $\mathcal{K}$, it outputs denotation~$y$.

\paragraph{Grounded Meaning Representation}

We represent grounded meaning representations in FunQL
\cite{kate2005learning} amongst many other alternatives such as lambda
calculus \cite{zettlemoyer_learning_2005}, $\lambda$-DCS
\cite{liang2013lambda} or graph queries
\cite{holzschuher2013performance,harris2013sparql}.  FunQL is a
variable-free query language, where each predicate is treated as a function symbol that modifies an argument list.
For example, the FunQL representation for the utterance
\textit{which states do not border texas} is:
\begin{myquote}{0.2in}
\T{answer}(\T{exclude}(\T{state(all)}, \T{next\_to}(\T{texas})))
\end{myquote}
where \T{next\_to} is a domain-specific binary predicate that takes one argument (i.e.,~the
entity \T{texas}) and returns a \textit{set} of entities (e.g.,~the
states bordering Texas) as its denotation.  \textit{all} is a special predicate that returns a collection of entities.  \T{exclude} is a predicate that returns the difference
between two input sets.

An advantage of FunQL is that the resulting
\mbox{\textit{s}-expression} encodes semantic compositionality and derivation of the logical
forms. This property makes FunQL logical forms natural to be generated with recurrent neural networks
\cite{vinyals2015grammar,charniakparsing,dyer2016recurrent}.  However, FunQL is less expressive than
lambda calculus, partially due to the elimination of variables.  
A more compact logical formulation which our method also applies to
is $\lambda$-DCS \cite{liang2013lambda}.
In the absence of anaphora and composite binary predicates, conversion
algorithms exist between FunQL and $\lambda$-DCS. However, we leave this
to future work.

\begin {table}[t]
\begin{center}
	\small
	\begin{tabular}{|@{~}L{1.6cm} L{3cm} L{2cm}@{~}|}
		\hline
		{Predicate} &  Usage & Sub-categories \\ \hline 
	   	 \T{answer}  & denotation wrapper  &\multicolumn{1}{c|}{---}   \\ \hline 
         \T{type} & entity type checking & \T{stateid}, \T{cityid}, \T{riverid}, etc.  \\  \hline
         \T{all} & querying for an entire set of entities & \multicolumn{1}{c|}{---} \\ \hline
         \T{aggregation} & one-argument meta predicates for sets  & \T{count}, \T{largest}, \T{smallest}, etc.  \\ \hline
         \T{logical connectives} & two-argument meta predicates for sets & \T{intersect}, \T{union}, \T{exclude} \\
		\hline
	\end{tabular}
\end{center}
\vspace{-2ex}
\caption{List of domain-general predicates.}
\label{general}
\end{table}

\paragraph{Ungrounded Meaning Representation}
We also use FunQL to express ungrounded meaning representations. The
latter consist primarily of natural language predicates and
domain-general predicates. Assuming for simplicity that domain-general
predicates share the same vocabulary in ungrounded and grounded
representations, the ungrounded representation for the example
utterance is:
\begin{myquote}{0.2in}
{\T{answer}(\T{exclude}(\T{states(all)}, \T{border}(\T{texas})))} 
\end{myquote}
where \T{states} and \T{border} are natural language predicates.  In
this work we consider five types of domain-general predicates
illustrated in Table~\ref{general}. Notice that domain-general
predicates are often implicit, or represent extra-sentential
knowledge.  For example, the predicate \T{all} in the above utterance
represents all states in the domain which are not mentioned in the
utterance but are critical for working out the utterance denotation. Finally, note that for certain domain-general predicates,
it also makes sense to extract natural language rationales (e.g., \textit{not} is indicative for \textit{exclude}).
But we do not find this helpful in experiments.



In this work we constrain ungrounded representations to be
structurally isomorphic to grounded ones.  In order to derive the
target logical forms, all we have to do is replacing predicates in the
ungrounded representations with symbols in the knowledge base.\footnote{As a more general definition, we consider two semantic graphs isomorphic
if the graph structures governed by domain-general predicates,  ignoring local structures containing only natural language predicates, are the same (Section \ref{sec:conclusions}). }



\section{Modeling}
\label{sec:modeling}

In this section, we discuss our neural model which maps utterances to target logical forms.  The semantic parsing task
is decomposed in two stages: we first explain how an utterance is converted
to an intermediate representation (Section~\ref{sec:gener-ungr-mean}),
and then describe how it is grounded to a knowledge base
(Section~\ref{sec:gener-ground-mean}).

\begin {table*}[t]
\begin{center}
	\begin{tabular}{  l | c | c | c}
		\multicolumn{4}{l}{	\textbf{Sentence}: \T{which states do not border texas} } \\
		\multicolumn{4}{l}{	\textbf{Non-terminal symbols in buffer}: \T{which}, \T{states}, \T{do}, \T{not}, \T{border} } \\
		\multicolumn{4}{l}{	\textbf{Terminal symbols in buffer}: \T{texas}} \\
		\textbf{Stack} & \textbf{Action} & NT choice & TER choice \\ \hline
		 &  \textsc{nt} & \T{\color{red} answer} &   \\
		\T{\color{red} answer} ( \quad\quad\quad\qquad\qquad\qquad\qquad\qquad\ & \textsc{nt} & \T{\color{red} exclude} &   \\
		\T{\color{red} answer} ( \T{\color{red} exclude} ( \qquad\qquad\qquad\qquad\qquad\,\;\,\; & \textsc{nt}  & \T{states}  &  \\
		\T{\color{red} answer} ( \T{\color{red} exclude}  ( \T{states} ( \quad\quad\qquad\qquad\quad\; &  \textsc{ter} &  &  {\color{red} \T{all}} \\
		\T{\color{red} answer} ( \T{\color{red} exclude}  ( \T{states} ( \T{\color{red} \T{all}} \quad\quad\qquad\qquad\quad\; & \textsc{red} &  & \\
		\T{\color{red} answer} ( \T{\color{red} exclude}  ( \T{states} ( \T{\color{red} \T{all}} ) \quad\quad\qquad\qquad\quad\; & \textsc{nt} & \T{border} & \\
		\T{\color{red} answer} ( \T{\color{red} exclude}  ( \T{states} ( \T{\color{red} \T{all}} ) , \T{border} (  \quad\quad\quad\:   & \textsc{ter} & & \T{texas} \\
		\T{\color{red} answer} ( \T{\color{red} exclude}  ( \T{states} ( \T{\color{red} \T{all}} ) , \T{border} ( \T{texas} \,\,\,\,\;  & \textsc{red} & & \\
		\T{\color{red} answer} ( \T{\color{red} exclude}  ( \T{states} ( \T{\color{red} \T{all}} ) , \T{border} ( \T{texas} ) \,\,\,\,\;  & \textsc{red} & & \\
		\T{\color{red} answer} ( \T{\color{red} exclude}  ( \T{states} ( \T{\color{red} \T{all}} ) , \T{border} ( \T{texas} ) ) & \textsc{red} & & \\
		\T{\color{red} answer} ( \T{\color{red} exclude}  ( \T{states} ( \T{\color{red} \T{all}} ) , \T{border} ( \T{texas} ) ) ) & & & \\
		\hline
	\end{tabular}
\end{center}
\caption{Actions taken by the transition system for
  generating the ungrounded meaning representation of the example utterance. Symbols in red indicate domain-general
  predicates. 
  \label{parser} }
\vspace{-2ex}
\end{table*}

\subsection{Generating Ungrounded Representations}
\label{sec:gener-ungr-mean}

At this stage, utterances are mapped to intermediate representations with
a transition-based algorithm.  
In general, the transition system generates the representation by following a derivation tree (which contains a set of applied rules) and 
some canonical generation order (e.g., pre-order).
For FunQL, a simple solution exists since the representation itself encodes the derivation.
Consider again \T{answer}(\T{exclude}(\T{states(all)},
\T{border}(\T{texas}))) which is tree structured.
Each predicate (e.g., \T{border}) can be visualized as a non-terminal
node of the tree and each entity (e.g.,~\T{texas}) as a terminal. The
predicate \T{all} is a special case which acts as a terminal directly.
We can generate the tree top-down with a transition system
reminiscent of recurrent neural network grammars (RNNGs;
\citealt{dyer2016recurrent}).  Similar to RNNG, our algorithm uses a
buffer to store input tokens in the utterance and a stack to store
partially completed trees.  A major difference in our semantic parsing
scenario is that tokens in the buffer are not fetched in a sequential
order or removed from the buffer.  This is because the lexical
alignment between an utterance and its semantic representation is
hidden.  Moreover, some domain-general predicates cannot be clearly anchored to a
token span.  Therefore, we allow the generation algorithm to pick
tokens and combine logical forms in arbitrary orders, conditioning on
the entire set of sentential features.  Alternative solutions in the
traditional semantic parsing literature include a floating chart
parser \cite{pasupat_compositional_2015} which allows to construct
logical predicates out of thin air.

Our transition system defines three actions, namely \textsc{nt},
\textsc{ter}, and \textsc{red}, explained below.

\paragraph{\textsc{nt(x)}} generates a \textsc{n}on-\textsc{t}erminal
predicate.  This predicate is either a natural language expression
such as \T{border}, or one of the domain-general predicates
exemplified in Table~\ref{general} (e.g.,~\T{exclude}). The type of
predicate is determined by the placeholder~\textsc{x} and once
generated, it is pushed onto the stack and represented as a
non-terminal followed by an open bracket (e.g.,~\T{`border('}).  The open bracket will be closed by a
reduce operation.

\paragraph{\textsc{ter(x)}} generates a \textsc{ter}minal entity or
the special predicate \textit{all}.  Note that the terminal choice
does not include variable (e.g.,~\$0, \$1), since FunQL is a
variable-free language which sufficiently captures the semantics of
the datasets we work with.  The framework could be extended to generate directed acyclic graphs by
incorporating variables with additional transition actions for handling
variable mentions and co-reference.

\paragraph{\textsc{red}} stands for \textsc{red}uce and is used for
subtree completion.  It recursively pops elements from the stack until
an open non-terminal node is encountered.  The non-terminal is popped
as well, after which a composite term representing the entire subtree,
e.g., \textit{border(texas)}, is pushed back to the stack.  If a
\textsc{red} action results in having no more open non-terminals left
on the stack, the transition system terminates.  Table~\ref{parser}
shows the transition actions used to generate our running example.

The model generates the ungrounded representation~$U$ conditioned on
utterance~$x$ by recursively calling one of the above three actions.
Note that $U$~is defined by a sequence of actions (denoted by~$a$) and
a sequence of term choices (denoted by~$u$) as shown in
Table~\ref{parser}.  The conditional probability $p(U|x)$ is
factorized over time steps as:
\begin{equation}
\begin{split}
p(U|x) &= p(a, u | x) \\
&= \prod_{t=1}^T p(a_t | a_{<t}, x) p(u_t | a_{<t}, x)^{\mathbb{I}(a_t \neq \textsc{red})} \\
\end{split}
\raisetag{3\baselineskip}
\end{equation}
where $\mathbb{I}$ is an indicator function.


To predict the actions of the transition system, we encode the input
buffer with a bidirectional LSTM \cite{hochreiter1997long} and the
output stack with a stack-LSTM \cite{dyer2015transition}.  At each
time step, the model uses the representation of the transition
system~$e_t$ to predict an action:
\begin{equation}
p(a_t | a_{<t}, x) \propto \exp (W_a \cdot e_t)
\end{equation}
where $e_t$ is the concatenation of the buffer representation $b_t$
and the stack representation~$s_t$.  While the stack
representation~$s_t$ is easy to retrieve as the top state of the
stack-LSTM, obtaining the buffer representation~$b_t$ is more
involved.  This is because we do not have an explicit buffer
representation due to the non-projectivity of semantic parsing.  We
therefore compute at each time step an adaptively weighted representation
of~$b_t$ \cite{bahdanau2014neural} conditioned
on the stack representation~$s_t$.  This buffer representation is then
concatenated with the stack representation to form the system
representation~$e_t$.

When the predicted action is either \textsc{nt} or \textsc{ter}, an
ungrounded term~$u_t$ (either a predicate or an entity) needs to be
chosen from the candidate list depending on the specific
placeholder~\textsc{x}.  To select a domain-general term, we use the
same representation of the transition system~$e_t$ to compute a
probability distribution over candidate terms:
\begin{equation}
p(u_t^{\textsc{GENERAL}} | a_{<t}, x) \propto \exp (W_p \cdot e_t)
\end{equation}
To choose a natural language term, we directly compute a probability distribution
of all natural language terms (in the buffer) conditioned on the
stack representation~$s_t$ and select the most relevant term
\cite{jia2016data,gu2016incorporating}:
\begin{equation}
p(u_t^{\textsc{NL}} | a_{<t}, x) \propto \exp (W_s \cdot s_t)
\end{equation}

When the predicted action is \textsc{red}, the completed subtree is
composed into a single representation on the stack.  For the choice of
composition function, we use a single-layer neural network as in
\newcite{dyer2015transition}, which takes as input the concatenated
representation of the predicate and arguments of the subtree.

\subsection{Generating Grounded Representations}
\label{sec:gener-ground-mean}

Since we constrain the network to learn ungrounded structures that are
isomorphic to the target meaning representation, converting ungrounded
representations to grounded ones becomes a simple lexical mapping
problem. For simplicity, hereafter we do not differentiate natural
language and domain-general predicates.

To map an ungrounded term~$u_t$ to a grounded term~$g_t$, we compute
the conditional probability of~$g_t$ given~$u_t$ with a bi-linear
neural network:
\begin{equation}
\label{eq:conditional}
p(g_t | u_t) \propto  \exp \vec{u_t} \cdot W_{ug} \cdot \vec{g_t}^\top 
\end{equation}
where $\vec{u_t}$ is the contextual representation of the ungrounded
term given by the bidirectional LSTM, $\vec{g_t}$~is the grounded term
embedding, and ${W}_{ug}$~is the weight matrix.

The above grounding step can be interpreted as learning a lexicon: the
model exclusively relies on the intermediate representation~$U$ to
predict the target meaning representation~$G$ without taking into
account any additional features based on the utterance. In
practice, $U$~may provide sufficient contextual background for closed
domain semantic parsing where an ungrounded predicate often maps to a
single grounded predicate, but is a relatively impoverished
representation for parsing large open-domain knowledge bases like
Freebase. In this case, we additionally rely on a discriminative
reranker which ranks the grounded representations derived from
ungrounded representations (see Section~\ref{sec:reranker}).

\subsection{Training Objective}
When the target meaning representation is available, we directly
compare it against our predictions and back-propagate.  When only
denotations are available, we compare surrogate meaning
representations against our predictions \cite{reddy2014large}.  Surrogate representations
are those with the correct denotations, filtered with rules (see Section~\ref{sec:experiments}).  When there exist multiple
surrogate representations,\footnote{The average Freebase surrogate representations obtained with highest denotation match (F1) is~1.4.}
we select one randomly and back-propagate.  


Consider utterance~$x$ with ungrounded meaning representation~$U$, and
grounded meaning representation~$G$. Both~$U$ and~$G$ are defined with
a sequence of transition actions (same for~$U$ and~$G$) and a sequence
of terms (different for~$U$ and~$G$).  Recall that $a = [a_1, \cdots,
a_n]$ denotes the transition action sequence defining~$U$ and~$G$; let
$u = [u_1, \cdots, u_k]$ denote the ungrounded terms
(e.g.,~predicates), and $g = [g_1, \cdots, g_k]$ the grounded
terms. We aim to maximize the likelihood of the grounded meaning
representation~$p(G | x)$ over all training examples.  This likelihood
can be decomposed into the likelihood of the grounded action
sequence~$p(a|x)$ and the grounded term sequence~$p(g|x)$, which we
optimize separately.

For the grounded action sequence (which by design is the same as the
ungrounded action sequence and therefore the output of the transition
system), we can directly maximize the log likelihood~$\log p(a | x)$
for all examples:
\begin{equation}
\mathcal{L}_a  = \sum\limits_{x \in \mathcal{T}}  \log p(a | x) 
=\sum\limits_{x \in \mathcal{T}}  \sum\limits_{t=1}^n \log p(a_t | x)
\end{equation}
where~$\mathcal{T}$ denotes examples in the training
data. 

For the grounded term sequence~$g$, since the intermediate ungrounded terms are
latent, we maximize the expected log likelihood of the grounded
terms $ \sum_{u} \left [ p(u | x) \log p(g | u, x) \right ]$ for all examples, which is a lower bound of the
log likelihood~$\log p(g | x)$ by Jensen's Inequality:
\begin{equation}
\begin{split}
\mathcal{L}_g & =\sum\limits_{x \in \mathcal{T}}   \sum_{u} \left [ p(u | x) \log p(g | u, x) \right ] \\
& =\sum\limits_{x \in \mathcal{T}}   \sum_{u} \left [ p(u | x) \sum\limits_{t=1}^k  \log p(g_t | u_t) \right ] \\
& \leq \sum\limits_{x \in \mathcal{T}}  \log p(g | x) 
\end{split}
\end{equation}

The final objective is the combination of $\mathcal{L}_a$
and~$\mathcal{L}_g$, denoted as $\mathcal{L}_G = \mathcal{L}_a +
\mathcal{L}_g$. We optimize this objective with the method described
in \newcite{lei2016rationalizing} and \newcite{xu2015show}.
%

\subsection{Reranker}
\label{sec:reranker}
As discussed above, for open domain semantic parsing, solely relying
on the ungrounded representation would result in an impoverished model
lacking sentential context useful for disambiguation decisions.  For all
Freebase experiments, we followed previous work
\cite{berant-EtAl:2013:EMNLP,berant2014semantic,reddy2014large} in
additionally training a discriminative ranker to re-rank grounded representations globally.

The discriminative ranker is a maximum-entropy model
\cite{berant-EtAl:2013:EMNLP}.  The objective is to maximize
the log likelihood of the correct answer $y$ given $x$ by summing over
all grounded candidates~$G$ with denotation~$y$ (i.e.,$[\![ G ]\!]_{\mathcal{K}} = y$):
\begin{equation}
\mathcal{L}_y  =\sum\limits_{(x, y) \in \mathcal{T}}  \log \sum\limits_{[\![ G ]\!]_{\mathcal{K}} = y} p (G | x)
\end{equation}
\begin{equation}
p (G | x) \propto \exp \{ f(G, x) \}  
\end{equation}
where~$f(G, x)$ is a feature function that maps pair~\mbox{($G$, $x$)}
into a feature vector. We give details on the features we used in
Section~\ref{sec:impl-deta}.


\section{Experiments}
\label{sec:experiments}

In this section, we verify empirically that our semantic parser
derives useful meaning representations. We give details on the
evaluation datasets and baselines used for comparison. We also
describe implementation details and the features used in the
discriminative ranker.

\subsection{Datasets}
\label{sec:datasets}
We evaluated our model on the following datasets which cover different
domains, and use different types of training data, i.e.,~pairs of
natural language utterances and grounded meanings or question-answer
pairs.

\textsc{GeoQuery} \cite{zelle1996learning} contains 880 questions and
database queries about US geography. The utterances are compositional,
but the language is simple and vocabulary size small. The majority of questions include at most one entity.
\textsc{Spades} \cite{bisk2016evaluating} contains 93,319 questions
derived from \textsc{clueweb09} \cite{gabrilovich2013facc1}
sentences. Specifically, the questions were created by randomly
removing an entity, thus producing sentence-denotation pairs
\cite{reddy2014large}.  The sentences include two or more entities and
although they are not very compositional, they constitute a
large-scale dataset for neural network training. \textsc{WebQuestions}
\cite{berant-EtAl:2013:EMNLP} contains~5,810 question-answer pairs.
Similar to \textsc{spades}, it is based on Freebase and the questions
are not very compositional. However, they are real questions asked by
people on the Web.
Finally, \textsc{GraphQuestions} \cite{su2016generating} contains
5,166 question-answer pairs which were created by showing 500~Freebase
graph queries to Amazon Mechanical Turk workers and asking them to
paraphrase them into natural language.

\subsection{Implementation Details}
\label{sec:impl-deta}

Amongst the four datasets described above, \textsc{GeoQuery} has
annotated logical forms which we directly use for training.  For the
other three datasets, we treat surrogate meaning representations which
lead to the correct answer as gold standard.  The surrogates were
selected from a subset of candidate Freebase graphs, which were
obtained by entity linking.  Entity mentions in \textsc{Spades} have
been automatically annotated with Freebase entities
\cite{gabrilovich2013facc1}.  For \textsc{WebQuestions} and
\textsc{GraphQuestions}, we follow the procedure described in
\newcite{reddy2016transforming}. We identify potential entity spans
using seven handcrafted part-of-speech patterns and associate them
with Freebase entities obtained from the Freebase/KG
API.\footnote{\url{http://developers.google.com/freebase/}} We use a
structured perceptron trained on the entities found in
\textsc{WebQuestions} and \textsc{GraphQuestions} to select the top~10
non-overlapping entity disambiguation possibilities.  We treat each
possibility as a candidate input utterance, and use the perceptron
score as a feature in the discriminative reranker, thus leaving the
final disambiguation to the semantic parser.

Apart from the entity score, the discriminative ranker uses the
following basic features.  The first feature is the likelihood score
of a grounded representation aggregating all intermediate
representations.  The second set of features include the embedding
similarity between the relation and the utterance, as well as the
similarity between the relation and the question words.  The last set
of features includes the answer type as indicated by the last word in
the Freebase relation \cite{xu2016question}.
 
We used the Adam optimizer for training with an initial learning rate
of~0.001, two momentum parameters [0.99, 0.999], and batch size~1.
The dimensions of the word embeddings, LSTM states, entity embeddings
and relation embeddings are $[50, 100, 100, 100]$.  The word
embeddings were initialized with Glove embeddings
\cite{pennington2014glove}.  All other embeddings were randomly
initialized.

\subsection{Results}

Experimental results on the four datasets are summarized in
Tables~\ref{webqa}--\ref{spade}.  We present comparisons of our system
which we call \textsc{ScanneR} (as a shorthand for
\textbf{S}ymboli\textbf{C} me\textbf{AN}i\textbf{N}g
r\textbf{E}p\textbf{R}esentation) against a variety
of models previously described in the literature. 

\geoq results are shown in Table~\ref{geo}. The first block contains
symbolic systems, whereas neural models are presented in the second
block.  We report accuracy which is defined as the proportion of the
utterance that are correctly parsed to their gold standard
logical forms. All previous neural systems \cite{dong2016language,jia2016data} treat semantic parsing as a
sequence transduction problem and use LSTMs to directly map utterances to logical forms. \textsc{ScanneR} yields performance improvements over these systems
 when using comparable data  sources
for training. \newcite{jia2016data} achieve better results with
synthetic data that expands \textsc{GeoQuery}; we could adopt their
approach to improve model performance, however, we leave this to
future work.

\begin {table}[t]
\begin{center}
	\small
	\begin{tabular}{|lc|}
		\hline
		{Models} & F1 \\ \hline\hline
		\newcite{berant-EtAl:2013:EMNLP} & 35.7\\
		\newcite{yao2014information} & 33.0 \\
		\newcite{berant2014semantic} & 39.9 \\
		\newcite{bast2015more} & 49.4 \\
		\newcite{berant2015imitation} & 49.7 \\
		\newcite{reddy2016transforming} & 50.3 \\ \hline \hline
		\newcite{bordesquestion} & 39.2 \\
		\newcite{dong2015question} & 40.8 \\
		\newcite{yih2015semantic} & 52.5 \\
		\newcite{xu2016question} & 53.3 \\
		Neural Baseline & 48.3 \\
		\textsc{ScanneR} & 49.4 \\
		\hline
	\end{tabular}
\end{center}
\vspace{-2ex}
\caption{\textsc{WebQuestions} results.}
\label{webqa}
\end{table}

\begin {table}[t!]
\begin{center}
	\small
	\begin{tabular}{|lr|}
          \hline
          {Models} & \multicolumn{1}{c|}{F1} \\ \hline \hline
          \textsc{sempre} \cite{berant-EtAl:2013:EMNLP} & 10.80 \\ %
          \textsc{parasempre} \cite{berant2014semantic} & 12.79 \\
          \textsc{jacana} \cite{yao2014information} & 5.08 \\
          Neural Baseline & 16.24 \\
          \textsc{ScanneR} & 17.02 \\
          \hline
	\end{tabular}
\vspace{-2ex}
\end{center}
\caption{\textsc{GraphQuestions} results. Numbers for comparison
  systems are from \newcite{su2016generating}.} 
\label{gqa}
\vspace{-2ex}
\end{table}

Table~\ref{spade} reports \textsc{ScanneR}'s performance on
\textsc{Spades}. For all Freebase related datasets we use average~F1
\cite{berant-EtAl:2013:EMNLP} as our evaluation metric.  Previous work
on this dataset has used a semantic parsing framework similar to ours
where natural language is converted to an intermediate syntactic
representation and then grounded to Freebase. Specifically,
\newcite{bisk2016evaluating} evaluate the effectiveness of four
different CCG parsers on the semantic parsing task when varying the
amount of supervision required. As can be seen, \textsc{ScanneR}
outperforms all CCG variants (from unsupervised to fully supervised)
without having access to any manually annotated derivations or
lexicons.  For fair comparison, we also built a neural baseline that
encodes an utterance with a recurrent neural network and then predicts a
grounded meaning representation directly
\cite{ture2016simple,yih2016value}.  Again, we observe that
\textsc{ScanneR} outperforms this baseline.

\begin {table}[t]
\begin{center}
	\small
	\begin{tabular}{|lc|}
		\hline
		{Models} & Accuracy \\ \hline \hline
		\newcite{zettlemoyer_learning_2005} & 79.3 \\ 
		\newcite{zettlemoyer2007online} & 86.1 \\
		\newcite{kwiatkowksi2010inducing} & 87.9\\
		\newcite{kwiatkowski2011lexical} & 88.6\\
		\newcite{kwiatkowski2013scaling} & 88.0\\
		\newcite{zhao2014type} & 88.9 \\
		\newcite{liang2011learning} & 91.1 \\
		\hline \hline
		\newcite{dong2016language}  & 84.6 \\
		\newcite{jia2016data} & 85.0 \\
		\newcite{jia2016data} with extra data & 89.1 \\
		\textsc{ScanneR}  & 86.7 \\
		\hline
	\end{tabular}
\end{center}
\vspace{-2ex}
\caption{\textsc{GeoQuery} results.}
\label{geo}
\end{table}


\begin {table}[t]
\begin{center}
	\small
	\begin{tabular}{|lc|}
		\hline 
		{Models} & F1 \\ \hline\hline
		Unsupervised CCG \cite{bisk2016evaluating} & 24.8 \\
		Semi-supervised CCG \cite{bisk2016evaluating} & 28.4 \\
		Neural baseline & 28.6 \\
		Supervised CCG \cite{bisk2016evaluating} & 30.9 \\
		Rule-based system \cite{bisk2016evaluating} & 31.4 \\
		\textsc{ScanneR} & 31.5 \\
		\hline
	\end{tabular}
\end{center}
\vspace{-2ex}
\caption{\textsc{Spades} results.}
\label{spade}
\end{table}

Results on \textsc{WebQuestions} are summarized in Table~\ref{webqa}.
\textsc{ScanneR} obtains performance on par with the best symbolic
systems (see the first block in the table). It is important to note
that \newcite{bast2015more} develop a question answering system, which
contrary to ours cannot produce meaning representations whereas
\newcite{berant2015imitation} propose a sophisticated agenda-based
parser which is trained borrowing ideas from imitation
learning. 
\newcite{reddy2016transforming} learns a semantic parser via
intermediate representations which they generate based on the output
of a dependency parser. \textsc{ScanneR} performs competitively
despite not having access to any linguistically-informed syntactic
structures. The second block in Table~\ref{webqa} reports the results
of several neural systems.  \newcite{xu2016question} represent the
state of the art on \textsc{WebQuestions}. Their system uses Wikipedia
to prune out erroneous candidate answers extracted from Freebase. Our
model would also benefit from a similar post-processing step. 
As in
previous experiments, \textsc{ScanneR} outperforms the neural baseline, too.

Finally, Table~\ref{gqa} presents our results on
\textsc{GraphQuestions}. We report F1 for \textsc{ScanneR}, the neural
baseline model, and three symbolic systems presented in
\newcite{su2016generating}. \textsc{ScanneR} achieves a new state of
the art on this dataset with a gain of~4.23 F1 points over the best
previously reported model.

\begin {table}[t]
\begin{center}
	\small
	\begin{tabular}{|lc|}
          \hline
          Metrics    & Accuracy\\ \hline \hline
          Exact match & 79.3 \\
          Structure match & 89.6 \\
          Token match & 96.5 \\
          \hline
	\end{tabular}
\end{center}
\vspace{-2ex}
\caption{\textsc{GeoQuery} evaluation of ungrounded meaning
  representations. We report accuracy against a manually created
  gold standard. }
\label{geoun}
\vspace{-2ex}
\end{table}


\subsection{Analysis of Intermediate Representations \label{analysisin}}
Since a central feature of our parser is that it learns intermediate
representations with natural language predicates, we conducted
additional experiments in order to inspect their quality.  For
\textsc{GeoQuery} which contains only 280 test examples, we manually
annotated intermediate representations for the test instances and
evaluated the learned representations against them.  
The experimental setup aims to show how humans can participate in improving
the semantic parser with feedback at the intermediate stage.
In terms of
evaluation, we use three metrics shown in Table~\ref{geoun}.  The
first row shows the percentage of exact matches between the predicted
representations and the human annotations.  The second row refers to
the percentage of structure matches, where the predicted
representations have the same structure as the human annotations, but
may not use the same lexical terms. Among structurally correct
predictions, we additionally compute how many tokens are correct, as
shown in the third row.  As can be seen, the induced meaning
representations overlap to a large extent with the human gold standard.

We also evaluated the intermediate representations created by
\textsc{ScanneR} on the other three (Freebase) datasets.  Since
creating a manual gold standard for these large datasets is
time-consuming, we compared the induced representations against the output of a syntactic
parser. Specifically, we converted the questions to event-argument
structures with \textsc{EasyCCG} \cite{lewis-steedman:2014:EMNLP2014},
a high coverage and high accuracy CCG parser. \textsc{EasyCCG}
extracts predicate-argument structures with a labeled F-score
of~83.37\%. For further comparison, we built a simple baseline which identifies predicates based on the
output of the Stanford POS-tagger \cite{manning2014stanford} following the
ordering VBD $\gg$ VBN $\gg$ VB $\gg$ VBP $\gg$ VBZ $\gg$ MD.

As shown in Table~\ref{spadeun}, on \textsc{Spades} and
\textsc{WebQuestions}, the predicates learned by our model match the
output of \textsc{EasyCCG} more closely than the heuristic baseline.
But for \graphq which contains more compositional questions, the
mismatch is higher.  However, since the key idea of our model is to
capture salient meaning for the task at hand rather than strictly obey
syntax, we would not expect the predicates induced by our system to
entirely agree with those produced by the syntactic parser.  To
further analyze how the learned predicates differ from syntax-based
ones, we grouped utterances in \textsc{Spades} into four types of
linguistic constructions: coordination (\T{conj}), control and raising
(\T{control}), prepositional phrase attachment (\T{pp}), and
subordinate clauses (\textit{subord}).  Table~\ref{spadeun} also shows the
breakdown of matching scores per linguistic construction, with the number of utterances in each type.  In
Table~\ref{example}, we provide examples of predicates identified by
\textsc{ScanneR}, indicating whether they agree or not with the output
of \textsc{EasyCCG}. As a reminder, the task in \textsc{Spades} is to
predict the entity masked by a \T{blank} symbol (\_\_).

\begin {table}[t]
\begin{center}
	\small
	\begin{tabular}{|l c c|}
		\hline
Dataset   & \textsc{ScanneR} & Baseline \\ \hline \hline
\textsc{Spades}     & 51.2 & 45.5  \\ 
\,\,\, --\textit{conj} (1422)  & 56.1 & 66.4 \\ 
\,\,\, --\textit{control} (132) &28.3 & 40.5\\
\,\,\, --\textit{pp} (3489) &46.2 & 23.1 \\ 
\,\,\, --\textit{subord} (76) &37.9 & 52.9 \\ \hline
\textsc{WebQuestions} & 42.1 & 25.5 \\ 
\textsc{GraphQuestions}  & 11.9 & 15.3 \\ \hline
\end{tabular}
\end{center}
\vspace{-2ex}
\caption{Evaluation of predicates induced by \textsc{ScanneR} against
  \textsc{EasyCCG}. We report F1(\%) across
  datasets. For \textsc{Spades}, we also provide a breakdown for
 various utterance types.}
\label{spadeun}
\vspace{-2ex}
\end{table}

\begin {table*}[t]
\begin{center}
	\small
	\begin{tabular}{|@{~}L{1.0cm}@{~} L{7cm}L{7.2cm}@{~}|}
		\hline
		\textit{conj} & 
		the boeing\_company was founded in 1916 and is \hlyellow{headquartered} in \_\_ , illinois .\newline
		nstar was founded in 1886 and is \hlyellow{based} in boston , \_\_ .\newline
		the \_\_ is owned and \hlyellow{operated} by zuffa\_,\_llc , headquarted in las\_vegas , nevada .\newline
		hugh \hlyellow{attended} \_\_ and then shifted to uppingham\_school in england .
		&
		\_\_ was \hlgreen{incorporated} in 1947 and is \hlgreen{based} \hlred{in} new\_york\_city . \newline
		the ifbb was \hlgreen{formed} in 1946 by president ben\_weider and his \hlred{brother} \_\_ . \newline
		wilhelm\_maybach and his \hlred{son} \_\_ \hlgreen{started} maybach in 1909 . \newline
		\_\_ was \hlgreen{founded} in 1996 and is \hlgreen{headquartered} \hlred{in} chicago . \\ \hline
		
		\textit{control} &
		\_\_ threatened to \hlyellow{kidnap} russ . \newline
		\_\_ has also been confirmed to \hlyellow{play} captain\_haddock . \newline
		hoffenberg decided to \hlyellow{leave} \_\_ . \newline
		\_\_ is reportedly trying to get \hlyellow{impregnated} by djimon now . \newline
		for right now , \_\_ are inclined to \hlyellow{trust} obama to do just that . 
		&
		\_\_ \hlred{agreed} to \hlgreen{purchase} wachovia\_corp . \newline
		ceo john\_thain \hlred{agreed} to \hlgreen{leave} \_\_ . \newline
		so nick \hlred{decided} to \hlgreen{create} \_\_ . \newline
		salva later \hlred{went} on to \hlgreen{make} the non clown-based horror \_\_ . \newline
		eddie \hlred{dumped} debbie to \hlgreen{marry} \_\_ when carrie was 2 . \\ \hline
				
		\textit{pp} & 
		\_\_ is the \hlyellow{home} of the university\_of\_tennessee .\newline
		chu is currently a physics \hlyellow{professor} at \_\_ .\newline
		youtube is \hlyellow{based} in \_\_ , near san\_francisco , california .\newline
		mathematica is a \hlyellow{product} of \_\_ .
		&
		jobs will \hlgreen{retire} \hlred{from} \_\_ .\newline
		the nab is a strong advocacy \hlgreen{group} \hlred{in} \_\_ .\newline
		this one \hlred{starred} robert\_reed , \hlgreen{known} mostly as \_\_ .\newline
		\_\_ is positively \hlgreen{frightening} as \hlred{detective} bud\_white .\\ \hline
		
		\textit{subord} &
		the\_\_ is a national testing board that is \hlyellow{based} in toronto .\newline
		\_\_ is a corporation that is wholly \hlyellow{owned} by the city\_of\_edmonton .\newline
		unborn is a scary movie that \hlyellow{stars} \_\_ .\newline
		\_\_ 's third \hlyellow{wife} was actress melina\_mercouri , who died in 1994 .\newline
		sure , there were \_\_ who \hlyellow{liked} the shah .
		&
		\hlgreen{founded} the \_\_ , which is now also a \hlred{designated} terrorist group .\newline
		\_\_ is an online \hlred{bank} that ebay \hlgreen{owns} .\newline
		zoya\_akhtar is a director , who has \hlgreen{directed} the upcoming \hlred{movie} \_\_ .\newline
		imelda\_staunton , who \hlgreen{plays} \_\_ , is \hlred{genius} .\newline
		\_\_ is the important \hlred{president} that american ever \hlgreen{had} .\newline
		plus mitt\_romney is the worst \hlred{governor} that \_\_ has  \hlgreen{had} . \\ 
		\hline
		
	\end{tabular}
\end{center}
\vspace{-2ex}
\caption{Informative predicates identified by \textsc{ScanneR} in various types of
  utterances. Yellow predicates were identified by both
  \textsc{ScanneR} and \textsc{EasyCCG}, red predicates by
  \textsc{ScanneR} alone, and green predicates by \textsc{EasyCCG}
  alone.}
\label{example}
\vspace{-2ex}
\end{table*}


As can be seen in Table~\ref{spadeun}, the matching score is
relatively high for utterances involving coordination and prepositional
phrase attachments.  The model will often identify informative
predicates (e.g., nouns) which do not necessarily agree with
linguistic intuition.  For example, in the utterance
\textit{wilhelm\_maybach and his son \_\_ started maybach in 1909}
(see Table~\ref{example}), \textsc{ScanneR} identifies the
predicate-argument structure \T{son(wilhelm\_maybach)} rather than
\T{started(wilhelm\_maybach)}.  We also observed that the model
struggles with control and subordinate constructions. It has
difficulty distinguishing control from raising predicates as
exemplified in the utterance \T{ceo john\_thain agreed to leave \_\_}
from Table~\ref{example}, where it identifies the control predicate
\T{agreed}. For subordinate clauses, \textsc{Scanner} tends to take
shortcuts identifying as predicates words closest to the \T{blank}
symbol.

\section{Discussion}
\label{sec:conclusions}
We presented a neural semantic parser which converts natural language
utterances to grounded meaning representations via intermediate
predicate-argument structures. Our model essentially jointly learns
how to parse natural language semantics and the lexicons that help
grounding.  Compared to previous neural semantic parsers, our model is
more interpretable as the intermediate structures are useful for
inspecting what the model has learned and whether it matches
linguistic intuition.

An assumption our model imposes is that ungrounded and grounded
representations are structurally isomorphic. An advantage of this
assumption is that tokens in the ungrounded and grounded
representations are strictly aligned.  This allows the neural network
to focus on parsing and lexical mapping, sidestepping the challenging
structure mapping problem which would result in a larger search space
and higher variance. On the negative side, the structural isomorphism
assumption restricts the expressiveness of the model, especially since
one of the main benefits of adopting a two-stage parser is the
potential of capturing domain-independent semantic information via the
intermediate representation.  While it would be challenging to handle
drastically non-isomorphic structures in the current model, it is
possible to perform local structure matching, i.e.,~when the mapping
between natural language and domain-specific predicates is many-to-one
or one-to-many.  For instance, Freebase does not contain a relation
representing \textit{daughter}, using instead two relations
representing \textit{female} and \textit{child}.  Previous work
\cite{kwiatkowski2013scaling} models such cases by introducing
collapsing (for many-to-one mapping) and expansion (for one-to-many
mapping) operators. Within our current framework, these two types of
structural mismatches can be handled with semi-Markov assumptions
\cite{sarawagi2004semi,kong2015segmental} in the parsing
(i.e.,~predicate selection) and the grounding steps, respectively.
Aside from relaxing strict isomorphism, we would also like to perform
cross-domain semantic parsing where the first stage of the semantic
parser is shared across domains.

\paragraph{Acknowledgments} We would like to thank three anonymous reviewers, members of the Edinburgh ILCC and the IBM Watson, and Abulhair Saparov for feedback. The support of the European Research Council
under award number 681760 ``Translating Multiple Modalities into
Text'' is gratefully acknowledged.

\bibliography{acl2017}
\bibliographystyle{acl_natbib}


\end{document}